%% file: camera_ready.tex
\documentclass[conference,a4paper]{IEEEtran}
\IEEEoverridecommandlockouts
\usepackage{cite}
\usepackage{amsmath,amssymb,amsfonts}
\usepackage{textcomp}
\usepackage{xcolor}
\usepackage{float}
\usepackage{epsfig}
\usepackage{graphicx}
\usepackage{multirow}
\usepackage[normalem]{ulem}
\usepackage{sidecap}
\usepackage{algorithm}
\usepackage{bbm}
\usepackage{algorithmicx}
\usepackage{algpseudocode}
\usepackage{booktabs}
\usepackage{tabularx} 
\usepackage{caption,subcaption} 
\usepackage{multicol}
\usepackage{url}
\usepackage{makecell}

\def\BibTeX{{\rm B\kern-.05em{\sc i\kern-.025em b}\kern-.08em
    T\kern-.1667em\lower.7ex\hbox{E}\kern-.125emX}}

\input{macros.tex}

\begin{document}

\title{ChatGPT-guided Semantics for Zero-shot Learning}
\author{
    \IEEEauthorblockN{
Fahimul Hoque Shubho\IEEEauthorrefmark{1}, 
Townim Faisal Chowdhury\IEEEauthorrefmark{2}, 
Ali Cheraghian\IEEEauthorrefmark{3}, 
\\Morteza Saberi\IEEEauthorrefmark{4}, 
Nabeel Mohammed\IEEEauthorrefmark{1}, 
Shafin Rahman\IEEEauthorrefmark{1}}
    \IEEEauthorblockA{
    \IEEEauthorrefmark{1}Dept. of Electrical and Computer Engineering, North South University, Bangladesh,\\
    \IEEEauthorrefmark{2}Australian Institute for Machine Learning, University of Adelaide, Australia,\\
    \IEEEauthorrefmark{3} Data61, Commonwealth Scientific and Industrial Research Organisation, Australia\\
    \IEEEauthorrefmark{4} School of Computer Science and DSI, University of Technology Sydney, Australia \\
    \{hoque.shubho, nabeel.mohammed, shafin.rahman\}@northsouth.edu,
    \\townim.chowdhury@adelaide.edu.au, ali.cheraghian@data61.csiro.au, 
    morteza.saberi@uts.edu.au}
}



\maketitle

\begin{abstract}
Zero-shot learning (ZSL) aims to classify objects that are not observed or seen during training. It relies on class semantic description to transfer knowledge from the seen classes to the unseen classes. Existing methods of obtaining class semantics include manual attributes or automatic word vectors from language models (like word2vec). We know attribute annotation is costly, whereas automatic word vectors are relatively noisy. To address this problem, we explore how ChatGPT, a large language model, can enhance class semantics for ZSL tasks. ChatGPT can be a helpful source to obtain text descriptions for each class containing related attributes and semantics. We use the word2vec model to get a word vector using the texts from ChatGPT. Then, we enrich word vectors by combining the word embeddings from class names and descriptions generated by ChatGPT. More specifically, we leverage ChatGPT to provide extra supervision for the class description, eventually benefiting ZSL models. We evaluate our approach on various 2D image (CUB and AwA) and 3D point cloud (ModelNet10, ModelNet40, and ScanObjectNN) datasets and show that it improves ZSL performance. Our work contributes to the ZSL literature by applying ChatGPT for class semantics enhancement and proposing a novel word vector fusion method.
\end{abstract}


\section{Introduction}

The goal of Zero-shot Learning (ZSL) is to classify unseen objects not observed in training. A more generic version termed Generalized ZSL (GZSL) attempts to predict a class from seen and unseen classes together. Researchers have started exploring ZSL and GZSL with 2D image datasets. Later, considering the availability of depth-sensing cameras, exploring ZSL on 3D point cloud data got considerable attention \cite{cheraghian2019zeroshot,cheraghian2021zero,zhang2021pointclip}. For both 2D and 3D cases, semantic descriptions of classes play a pivotal role in transferring knowledge from seen to unseen classes. Class semantics are designed to describe all objects with a common set of features or components, working as a bridge between seen and unseen worlds. The literature shows that addressing ZSL tasks in 3D has more challenges than its 2D counterpart. Therefore, improving class semantics may help to address some challenges. This paper attempts to improve class semantic descriptions for 2D and 3D ZSL tasks.

\begin{figure}[!t]
\centering
\begin{center}
\includegraphics[width=1\columnwidth]{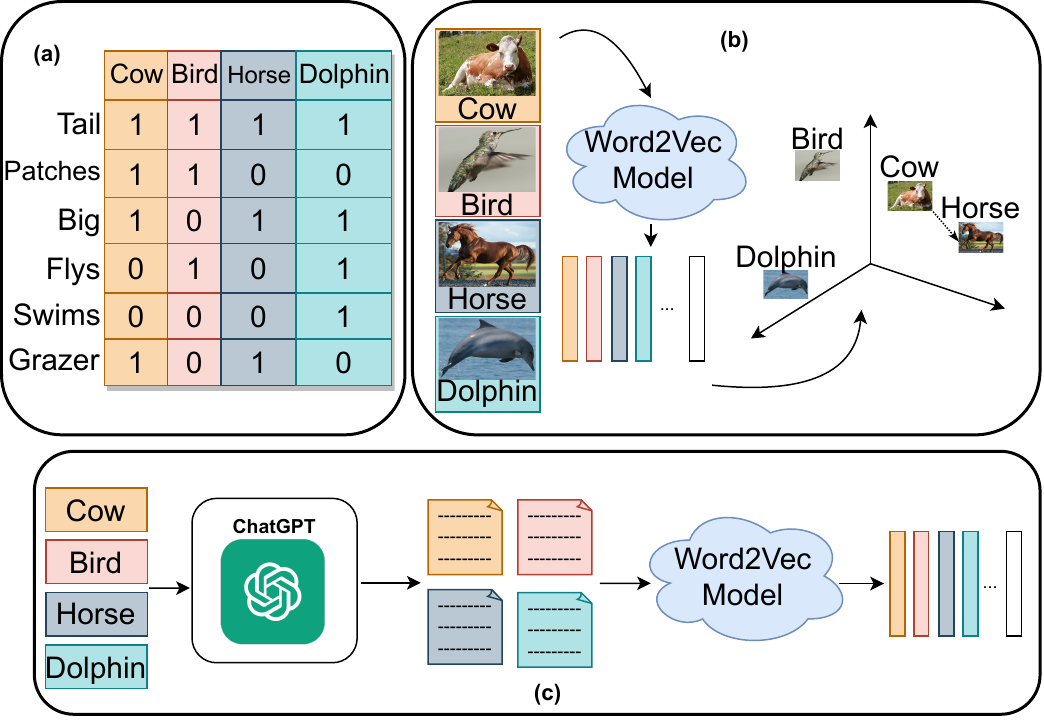}  
\end{center}
\caption{Class semantics can be obtained from (a) manually annotated attributes or (b) automatically calculated from language models like word2vec. Attributes need hard manual labor, whereas word vectors are relatively noisy. In this paper (c), we attempt the improve the quality of automatic word vectors by using extra supervision from ChatGPT.}
\label{fig:intro}
\end{figure}

Class semantics can be obtained manually (attribute vectors) or automatically (word vectors). Attributes are identifiable features to describe a class (see Fig. \ref{fig:intro}(a)) that require laborious human annotation to obtain and are not readily available for many large-scale or 3D point cloud datasets. In contrast, automatic word vectors are the output (as vectors) of language models (like word2vec \cite{mikolov2013efficient}, GloVe \cite{pennington-etal-2014-glove}, Fasttext \cite{bojanowski2016enriching}, Bert \cite{devlin2019bert}, etc.), given class names as input  (see Fig. \ref{fig:intro}(b)). These models are usually trained using billions of text corpus from Wikipedia, news articles, etc. Compared to attributes, the automatic extraction of word vectors makes them more realistic for real-world applications. However, embeddings from word vectors are noisier than manual attribute vectors resulting in poorer ZSL performance than attributes. This issue becomes more challenging, especially for ZSL on 3D point cloud objects, because of pre-trained models, poor quality features, dataset size, etc. \cite{cheraghian2021zero}. In this paper, we investigate the use of ChatGPT \cite{brown2020language} to improve class semantics.

Usually, word vectors are calculated using a single class name as input. However, many related words and definitions associated with that single class name are also necessary to improve the representativeness of the word vector. Considering related semantics can improve the semantic description of the given class \cite{Rahman_TNNLS_2022}. Again, such related semantics can be obtained by hard manual annotations \cite{Zhang_2017_CVPR}, a costly process, or noisy web crawling annotation~\cite{radford2021learning}. To address this problem, recent ChatGPT can be a valuable source for describing a class with related semantics and attributes (see Fig. \ref{fig:intro}(c)). It is both automatic and less noisy. Therefore, given a class name, we ask ChatGPT to describe that class with a paragraph of text containing related semantics and attributes. Then, we extract a word embedding of that paragraph using a language model(word2vec). Finally, we linearly combine word embeddings from class names and ChatGPT to calculate an improved word vector. Different from \cite{zhang2021pointclip}, this approach does not need any prompt engineering, so it can be used in any existing ZSL models to increase accuracy. We test our approach on seven methods DEM \cite{Zhang_2017_CVPR}, LATEM \cite{latem-cvpr16}, SYNC \cite{Changpinyo_2016_CVPR}, GDAN \cite{gdan-cvpr19}, f-CLSWGAN \cite{Xian_2018_CVPR}, TF-VAEGAN \cite{tfvaegan-eccv20} and CADA-VAE \cite{Schonfeld_2019_CVPR} covering both 2D image (CUB \cite{WahCUB_200_2011} and AwA \cite{Xian_CVPR_2017}) and 3D point cloud datasets (ModelNet10, ModelNet40 \cite{wu20153d} and ScanObjectNN \cite{uy2019revisiting}). Note that our method is also applicable for both synthetic ModelNet40 and real-world scanned ScanObjectNN cases. We consistently improve performance in experiments across different ZSL and GZSL problem setups.

Our contributions are as follows:
\begin{itemize}
    \item Employing ChatGPT guided semantics for solving ZSL problems,

    \item Proposing an approach that uses texts provided by ChatGPT as extra supervision to enhance word vectors while representing class semantics that readily be plugged into any ZSL method to improve accuracy,

    \item Extensive experiments on 2D images and 3D point cloud object (synthetic and real-world scanned) datasets.
\end{itemize}

\section{Related works}

\noindent\textbf{ZSL with 2D images:} Initial work of ZSL~\cite{6571196} used image-related information attributes to classify images in the unseen test set, and the model was trained by utilizing the seen attributes. In another work, \cite{frome2013devise} proposed using labeled image data and semantic information gleaned from unlabeled text using a deep visual-semantic embedding model. The model learns the semantic relationship between labels from the textual data and directly maps images into a deep semantic embedding space. In a later work, Akata et al. \cite{Akata_CVPR_2015} used image features with supervised attributes and unsupervised output embedding derived from unlabeled text corpora to project image features in attribute space and measure similarity with class description to classify images. In \cite{Kodirov_2015_ICCV}, Kodirov et al. proposed learning in the reverse direction of the previous work by using an unsupervised domain adaption approach that employs the target domain class labels’ projections in the semantic space to regularise the learned target domain projection. In contrast to using a global linear embedding like previous works, \cite{latem-cvpr16} learns various linear models and allows each image-class combination to be chosen from them. Recent works use a generative model-based approach to mitigate bias and domain shift problems observed in the ZSL. In \cite{mishra2018generative}, Mishra et al. proposed procreating samples using the attributes with conditional variational autoencoder and used the created samples to classify the unseen classes. In \cite{lsrgan-eccv20}, the authors leveraged the semantic relationship between the seen and unseen classes to generate visual features for the unseen courses while maintaining the same semantic relationship between both classes. Hayat et al. \cite{10.1007/978-3-030-69535-4_10} proposed synthesizing the features for unseen classes using a generative model with their class semantics for ZSL in object detection. This paper aims to ZSL and GZSL tasks based on improved class semantics.

\noindent\textbf{ZSL with 3D point cloud object:} The works around 3D ZSL are relatively new compared to 2D images.  In a pioneer work, Cheraghian et al. \cite{cheraghian2019zero} proposed using 3D point clouds objects for zero-shot tasks following an inductive process with transductive ZSL to classify 3D cloud points and show that it can achieve competitive performance for 3D point cloud classification. 
Later, \cite{cheraghian2019mitigating} proposed a loss function comprised of a regression and skewness term to reduce the hubness problems described in \cite{cheraghian2019zero}. The authors used transductive ZSL and Generalized Zero-Shot Learning (GZSL) approaches for classifying 3D point data. They also introduced a triplet loss function for these approaches. \cite{cheraghian2021zero} proposed a unified approach to address the hubness and domain shift problems using a novel loss function. \cite{9665941} introduced a generative approach to classify and segment 3D data for ZSL and GZSL. In \cite{10034623}, the authors used NLP based prompt to augment new data that improves the performance of ZSL on 3D datasets. This paper investigates the performance of both embedding and generative methods using improved class semantics.

\noindent\textbf{LLM in computer vision tasks:} Large Language Models (LLM) like GPT have been used widely in natural language processing tasks. Recent works have used LLM for computer vision tasks also. \cite{Zhao_2023_CVPR} leveraged LLM to generate narration from video input. \cite{he2023annollm} proposed that LLM models such as GPT-3.5 can annotate unlabeled data. They also showed that the annotation from LLM works better for user input and keyword relevance assessment compared to using crowdsourced annotation data. Later, \cite{yu2023using} used 2 LLM, GPT-3.5 and GPT-4, for automatic annotation of texts with specific categories of linguistic information and compared it with human annotator to show that it is feasible to use LLM with the annotation for local grammar analysis. In \cite{whitehouse2023llmpowered}, the authors used several LLMs, such as Dolly-v2, StableVicuna, ChatGPT, and GPT-4, for data augmentation in cross-lingual commonsense reasoning datasets to overcome the imitation of training data and showed that it has a positive impact on the overall performance of the trained models. \cite{NEURIPS2022_0346c148} applied a language model to procreate class-conditioned texts attuned by prompts and utilized them as the training data for fine-tuning a pre-trained bidirectional language model used for zero-shot learning of Natural Language Understanding tasks. This paper uses ChatGPT to ensure extra supervision for class semantics during ZSL and GZSL tasks.

\section{Method}



\begin{figure}[!t]
\centering
\begin{center}
\includegraphics[width=.5\textwidth]{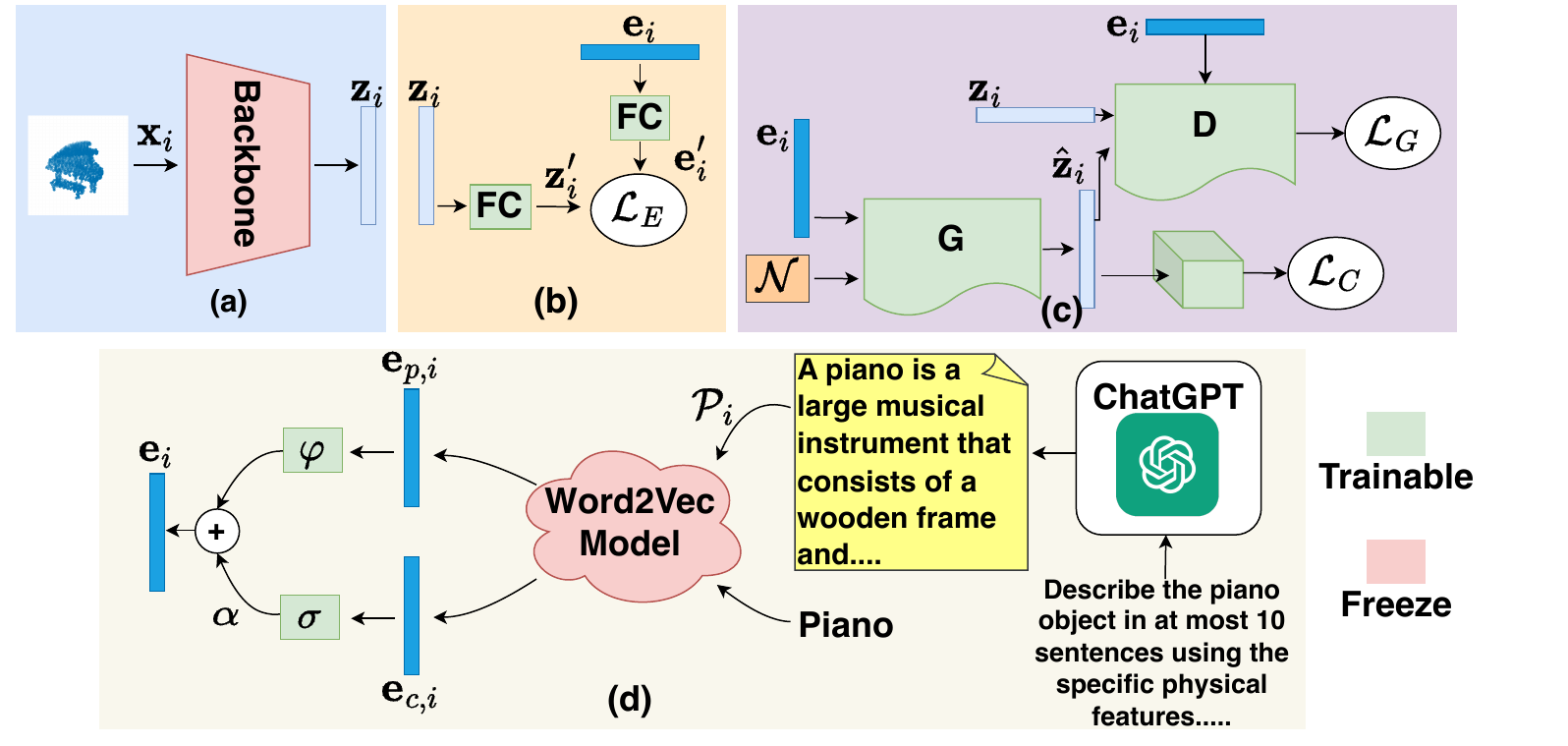}  
\end{center}

\small\caption{We demonstrate the profound impact of integrating ChatGPT-guided semantic information into traditional ZSL techniques. (a) The input data $\textbf{x}_{i} \in \mathbb{R}^K$ (e.g., images or point clouds) is initially processed through a Backbone, producing a feature embedding $\textbf{z}_{i} \in \mathbb{R}^m$. (b) ZSL methods often utilize a shared common space, where the distance between the mapped feature embeddings, $\textbf{z}_{i} \in \mathbb{R}^m$ and $\textbf{e}_{i} \in \mathbb{R}^d$, is minimized through a loss function. (c) In generative ZSL methods, a generator model receives a semantic embedding $\textbf{e}_{i} \in \mathbb{R}^d$, along with Gaussian noise, as input to produce synthetic feature embeddings $\hat{\textbf{z}}_{i} \in \mathbb{R}^{m}$. (d) In our proposed method, we leverage ChatGPT to generate additional descriptions for classes belonging to the seen category. Both the ChatGPT output and the class names are fed into a Word2vec model to extract their respective semantic feature embeddings. These extracted feature embeddings are then passed through fully connected layers and merged in the subsequent stage to generate the final comprehensive semantic feature embeddings, $\textbf{e}_{i} \in \mathbb{R}^{d}$.}
\label{fig_1}
\end{figure}

We demonstrate the potential of leveraging the generation capability of LLMs to enhance ZSL methods for recognizing unseen classes. LLMs have gained significant attention due to numerous advantages across various applications. To harness this power, we specifically utilize the ChatGPT model to generate supplementary descriptions for the class names during the training phase of ZSL. This description includes attributes and semantics related to a class that can enhance its discriminative ability from other classes. Encoding this description with a word vector by forwarding it through a language model (word2vec) can augment additional supervision to ZSL models.

\subsection{Problem formulation}

Let $\textbf{x} \in \mathbb{R}^K$ represent the input data, corresponding to either an image or point cloud data. We have two sets of class labels, $\sY^{s} = \{y_{1}^{s},...,y_{S}^{s}\}$ for seen classes and $\sY^{u} = \{y_{1}^{u},...,y_{U}^{u}\}$ for unseen classes. Here, the seen and unseen labels are disjoint, i.e., $\sY^{s}\cap\sY^{u}=\emptyset$. Additionally, $\sE^{s}=\{\phi(y_{1}^{s}), ..., \phi(y_{S}^{s})\}$ and $\sE^{u}=\{\phi(y_{1}^{u}), ..., \phi(y_{U}^{u})\}$ represent the sets of semantic feature embeddings obtained using the embedding function $\phi(\cdot)$, where $\phi(y)\in\bbR^{d}$. To proceed, we define the set of $n_{s}$ seen samples as $\mathcal{D}^{s} = \{(\textbf{x}_{i}^{s}, y_{i}^{s}, \textbf{e}_{i}^{s})\}_{i=1}^{n_{s}}$, where $\textbf{x}_{i}^{s}$ represents the $i$\textsuperscript{th} instance from the seen set, with ground truth label $y_{i}^{s} \in \sY^{s}$ and corresponding semantic vector $\be_{i}^{s} = \phi(y_{i}^{s}) \in \sE^{s}$. Similarly, the set of $n_{u}$ unseen samples is defined as $\mathcal{D}^{u} = \{(\textbf{x}_{i}^{u}, y_{i}^{u}, \textbf{e}_{i}^{u})\}_{i=1}^{n_{u}}$, where $\textbf{x}_{i}^{u}$ represents the $i$\textsuperscript{th} sample from the unseen set, with ground truth label $y_{i}^{u} \in \sY^{u}$ and corresponding semantic vector $\be_{i}^{u} = \phi(y_{i}^{u}) \in \sE^{u}$. This paper addresses two main tasks: Zero-Shot Learning (ZSL) and Generalized Zero-Shot Learning (GZSL). In ZSL, we assign unseen class labels $\hat{\textbf{y}} \in \mathcal{Y}^{u}$ to unseen instances $\textbf{x} \in \mathcal{D}^{u}$. In GZSL, we assign class labels $\hat{\textbf{y}} \in \mathcal{Y}^{s} \cup \mathcal{Y}^{u}$ to instances $\mathbf{x} \in \mathcal{D}^{s} \cup \mathcal{D}^{u}$, encompassing both seen and unseen classes. The semantic vectors $\be_{i}^{s}$ and $\be_{i}^{u}$ are usually implemented with attribute sets or word vectors. This paper proposes a method to produce an improved version of word2vec-based word vectors.





\begin{table*}[!t] \centering \small

\scalebox{.77}{
\begin{tabular}{ l|l|l|l  } 
\hline
\multirow{2}{4.5em}{Input Prompt} & \multicolumn{3}{l}{Describe the \textcolor{red}{piano} object in at most ten sentences using the specific physical features and do not need to mention the features that are not available in the object.} \\
& \multicolumn{3}{l}{Also, do not use any numeric in descriptions; instead, use words.}\\ 

\hline
\multirow{5}{4.5em}{ChatGPT Generate Description} & \multicolumn{3}{l}{A piano is a large musical instrument that consists of a wooden frame and various physical components.} \\
&\multicolumn{3}{l}{The main body of the piano contains the strings and soundboard.}\\
&\multicolumn{3}{l}{The strings are stretched horizontally across the length of the piano, with different lengths and thicknesses producing various pitches when struck.}\\
&\multicolumn{3}{l}{The soundboard is a large wooden panel located underneath the strings, responsible for resonating and amplifying the vibrations produced by the strings.}\\
&\multicolumn{3}{l}{The piano has a keyboard consisting of black and white keys.}\\
\hline

\end{tabular}}
\small\caption{A sample input prompt to ChatGPT and generated text description. The word vector generated from the text description of a class name (\textcolor{red}{piano}) can encode auxiliary semantics inside the word vector of \textcolor{red}{piano}.}
\label{tab:ChatGPT_generated_description}
\end{table*}

\begin{figure*}[!t]
\centering
\includegraphics[width=.9\textwidth]{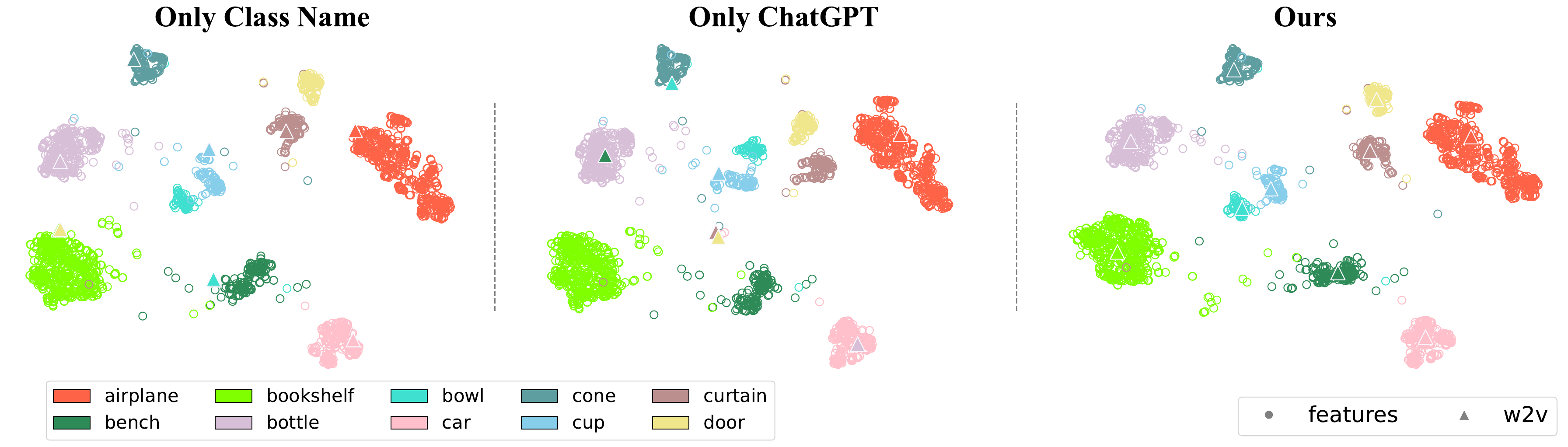}  
\caption{tSNE visualization of features and semantics of 10 classes from ModelNet40 dataset. Our method has achieved better feature-semantic alignment than only Class name and ChatGPT-based embedding vectors.}
\label{fig_tSNE_visualization}
\end{figure*}

\subsection{Preliminaries}

Two main types ZSL methods are embedding~\cite{wu20153d, uy2019revisiting, 6571196, frome2013devise, Akata_CVPR_2015} and generative~\cite{Article62, mishra2018generative, 10.1007/978-3-030-69535-4_10}. Embedding methods learn to map both the visual features of the input data and the semantic attributes to a common space. They can recognize unseen classes in this space by comparing their attribute representations with the input's features. In contrast, generative methods synthesize samples that look like unseen classes by using a mix of labeled data from seen classes and auxiliary information, such as class attributes or textual descriptions. 

Overall, in any ZSL methodology, an input backbone transfers the input data $\textbf{x}_{i} \in \mathbb{R}^K$ into a meaningful feature embedding, denoted as $\textbf{z}_{i} \in \mathbb{R}^{m}$, see Fig~\ref{fig_1} (a). Next, we will delve deeper into the embedding and synthetic Zero-Shot Learning (ZSL) methods.

\noindent\textbf{Embedding ZSL:} Embedding-based ZSL aims to learn functions that map input data (images or point clouds) and semantic information (attributes or class labels) to a shared space. In this space, the model can link input features with semantic information, allowing it to identify and categorize classes it has not seen before. In general, There are two branches in the embedding approach Fig~\ref{fig_1} (b). In the first branch, the input feature embedding $\textbf{z}_{i} \in \mathbb{R}^{m}$ is forwarded into a fully connected layer, resulting in $\textbf{z}'_{i} \in \mathbb{R}^{q}$ . Simultaneously, the corresponding semantic representation $\textbf{e}_{i} \in \mathbb{R}^{d}$ is fed into a fully connected layer, which generates $\textbf{e}'_{i} \in \mathbb{R}^{q}$ to map the semantic embedding to a common space, where the Euclidean distance between the semantic and feature embeddings is minimized. This is achieved by optimizing the following objective function:

\begin{equation}
\begin{aligned}
\mathcal{L}_{E} &= \frac{1}{n_{s}} \sum_{i=1}^{n_{s}} \left\| \textbf{z}'_{i} - \textbf{e}'_{i} \right\|_{2}^{2} + \lambda \mathcal{R}(\theta)
\end{aligned}
\end{equation}
where the $\theta$ is the trainable parameters, and $\mathcal{R}$ refers to the regularization loss function. The hyperparameter $\lambda$ is crucial in controlling the trade-off between the regularization and embedding losses.


\noindent\textbf{Generative ZSL:} In generative Zero-Shot Learning (ZSL) methods~\cite{gdan-cvpr19, Xian_2018_CVPR}, the objective is to generate synthetic samples for the unseen classes based on their semantic attributes. These methods enable the model to generate realistic and representative samples of unseen classes by leveraging the semantic information associated with those classes. 

To be more specific, our goal is to learn a conditional Generative Adversarial Network (GAN) model, denoted as ${G}$, which takes random Gaussian noise $\textbf{h} \sim \mathcal{N}(0,1)$ and the semantic class embedding $\textbf{e}_{i}$ as inputs to generate the feature representation of class $i$, denoted as $\hat{\textbf{z}}_{i} \in \mathbb{R}^{m}$. Concurrently, we train a discriminator model, $D$, to classify real features ${\textbf{z}}_{i} \in \mathbb{R}^{m}$ against synthetic features $\hat{\textbf{z}}_{i} \in \mathbb{R}^{m}$.

The objective function for generating synthetic feature samples is defined as follows:

\begin{equation}
\begin{aligned}
\mathcal{L}_{G} = & \mathbb{E}_{\textbf{z}, \textbf{e}}[D(\textbf{z}, \textbf{e})] - \mathbb{E}_{\textbf{z}, \textbf{e}}[D(\hat{\textbf{z}}, \textbf{e})] \\
& - \eta \mathbb{E}_{\tilde{\textbf{z}}, \textbf{e}}\left[\left(\left\|\nabla_{\tilde{\textbf{z}}} D(\tilde{\textbf{z}}, \textbf{e})\right\|_2 - 1\right)^2\right],
\end{aligned}
\end{equation}

where $\tilde{\textbf{z}} = \beta\textbf{z} + (1-\beta)\hat{\textbf{z}}$, with $\beta \sim \mathcal{U}(0,1)$, and $\eta$ is the penalty coefficient. This objective function guides the training of our conditional GAN model, enabling it to generate realistic and diverse feature representations based on the input class embeddings and Gaussian noise.





Also, a classification loss is required to ensure that the synthetic samples are suitable for the classifier. This loss is defined as follows:
\begin{equation}
\mathcal{L}_{C}=-E_{\hat{\textbf{z}} \sim p_{\hat{\textbf{z}}}}[\log P(\textbf{y} \mid \hat{\textbf{z}} ; \Theta)],
\end{equation}

The provided loss function is computed using a linear softmax classifier parameterized by $\Theta$, which undergoes pre-training on the real features $\textbf{z} \in \mathcal{D}^{s}$ from seen classes. To elaborate, this loss function serves as a regularizer, motivating the generator to construct discerning features in its generated samples.

\subsection{Improved class semantics}



We utilize the ChatGPT model by providing it with the class name of a seen class, $\textbf{y}^{s} \in \mathcal{Y}^{s}$, to generate descriptive sentences. Specifically, we input the class names into the GPT-3.5~\cite{brown2020language}, using prompts such as ``Describe the [CLASS] in at most ten sentences, focusing on specific physical features and excluding any unavailable features." The generated sentences offer detailed descriptions of the class names, aiding in improving class semantics for our ZSL tasks.
We denote the prompts created for each sample, formulated as follows:
\begin{equation}
\mathcal{P} = \text { ChatGPT(Commands). }
\end{equation}
An illustrative example of a class name description obtained from ChatGPT is presented in the following Table~\ref{tab:ChatGPT_generated_description}.

\begin{table*}[!t] \centering \small
\newcolumntype{C}{>{\centering\arraybackslash}X}
\setlength{\tabcolsep}{4pt}
\scalebox{.9}{
\begin{tabularx}{\textwidth}{clccCCCCC|CCCCC} \hline
\multirow{3}{*}{} & \multirow{3}{*}{Method } & & & \multicolumn{5}{c}{\textbf{AwA}} & \multicolumn{5}{c}{\textbf{CUB}}\\
\cline{5-14}
{} & {} & {} & {} & ZSL & \multicolumn{4}{c|}{GZSL} & ZSL & \multicolumn{4}{c}{GZSL} \\
 & & & Variations & Acc & $\Acc_{s}$ & $\Acc_{u}$ & HM & BC  & Acc & $\Acc_{s}$ & $\Acc_{u}$ & HM & BC  \\ 
 \hline
& \multirow{3}{*}{DEM \cite{Zhang_2017_CVPR}} & & Only Class Name & 45.79 & 83.15 & 10.95  & 19.35 & 0 & 10.68 & 18.47 & 2.62 & 4.59 & 0 \\ 
& & & Only ChatGPT & \textbf{54.74} & 56.09 & 8.71 & 15.09 & 1 & 11.69 & \textbf{47.78} & 1.62 & 3.14 & 1 \\
& & & Ours & 52.26 & \textbf{85.45} & \textbf{15.33} & \textbf{25.99} & \textbf{3} & \textbf{15.17} & 30.91 & \textbf{2.63} & \textbf{4.85} & \textbf{3} \\
\hline
& \multirow{3}{*}{LATEM \cite{latem-cvpr16}} & & Only Class Name & 46.21 &  - & - & - & 0 & 11.35 & - & - & - & 0\\ 
& & & Only ChatGPT & 50.47 &  - & - & - & 0 & 10.19 & - & - & - & 0 \\
& & & Ours & \textbf{52.31} &  - & - & - & \textbf{1} & \textbf{14.50} & - & - & - & \textbf{1} \\
\hline
& \multirow{3}{*}{SYNC \cite{Changpinyo_2016_CVPR}} & & Only Class Name & 48.83 &   - & - & - & 0 & 12.71 & - & - & - & 0 \\ 
& & & Only ChatGPT & \textbf{55.99} & - & - & - & \textbf{1} & 11.90 & - & - & - & 0 \\
& & & Ours & 55.10 &  - & - & - & 0 & \textbf{15.28} & - & - & - & \textbf{1} \\ 
\hline
& \multirow{3}{*}{GDAN \cite{gdan-cvpr19}} & & Only Class Name & - & \textbf{60.07} & 7.88 & 13.93 & 1	& - & 21.02 & \textbf{6.54} & \textbf{9.97} & \textbf{2} \\ 
& & & Only ChatGPT & - & 56.92 & 20.29 & 29.91 & 0 & - & \textbf{27.09} & 1.12 & 2.15 & 1 \\
& & & Ours & - & 56.35 & \textbf{23.48} & \textbf{33.15} & \textbf{2} & - & 20.72 & 6.44 & 9.83 & 0 \\
\hline
& \multirow{3}{*}{f-CLSWGAN \cite{Xian_2018_CVPR}} & & Only Class Name & 42.48 & \textbf{91.4} & 0.62 & 1.24 & 1 & 12.56 & \textbf{59.01} & 2.00 & 3.87 & 1 \\ 
& & & Only ChatGPT & 45.05 & 90.81 & 0.71 & 1.41 & 0 & 11.84 & 58.34 & 1.64 & 3.19 & 0 \\
& & & Ours & \textbf{45.74} & 91.00 & \textbf{0.76} & \textbf{1.50} & \textbf{3} & \textbf{15.37} & 57.41 & \textbf{3.20} & \textbf{6.05} & \textbf{3} \\
\hline
& \multirow{3}{*}{TF-VAEGAN \cite{tfvaegan-eccv20}} & & Only Class Name & 56.3 & 72.44 & \textbf{45.34} & 55.37 & 1 & 16.55 & 30.77 & 14.01 & 19.25 & 0 \\ 
& & & Only ChatGPT & \textbf{64.31} & \textbf{78.72} & 40.53 & 53.51 & \textbf{2} & 15.51 & 26.51 & \textbf{14.68} & 18.89 & 1 \\
& & & Ours & 57.14 & 73.67 & 44.54 & \textbf{55.47} & 1 & \textbf{20.46} & \textbf{48.39} & 14.64 & \textbf{22.48} & \textbf{3} \\
\hline
& \multirow{3}{*}{CADA-VAE \cite{Schonfeld_2019_CVPR}} & & Only Class Name & 36.79 & \textbf{90.02} & 19.16 & 31.6 & 1 & 16.17 & 64.71 & 4.65 & 8.68 & 0 \\ 
& & & Only ChatGPT & \textbf{45.01} & 82.93 & 20.51 & 32.89 & 1 & 11.71 & 63.44 & 4.02 & 7.56 & 0 \\
& & & Ours & 40.17 & 89.25 & \textbf{20.69} & \textbf{33.59} & \textbf{2} & \textbf{16.50} & \textbf{64.97} & \textbf{5.30} & \textbf{9.79} & \textbf{4} \\
\hline
\end{tabularx}}
\vspace{.2em}
\caption{ZSL and GZSL results on 2D image datasets.}
\label{tab:2D_results}
\end{table*}

In our innovative approach, visualized in Fig.~\ref{fig_1} (d), we have devised a multi-step process to enrich our understanding of class characteristics. The class name is initially processed by ChatGPT, generating a comprehensive description denoted as $\mathcal{P}_{i}$. This description includes multiple sentences closely related to the class name, as discussed earlier.

Subsequently, both the class name's semantic representation, denoted as $\textbf{e}_{c,i}\in \mathbb{R}^{d}$, and the ChatGPT output's semantic representation, denoted as $\textbf{e}_{p,i}\in \mathbb{R}^{d}$, are extracted using a Word2vec model. These semantic representations offer valuable insights into the meaning and context of the class name and its associated description. Afterwards, these semantic descriptions are passed through separate fully connected layers, represented by $\sigma$ and $\alpha$ respectively, before being merged using the following equation:
\begin{equation}
\mathbf{e}_{i} = \sigma(\textbf{e}_{c,i}) + \alpha \cdot \varphi(\textbf{e}_{p,i})
\end{equation}


The merging process through addition allows the combination of the class-specific information with the contextual details from ChatGPT, resulting in a more comprehensive and enriched representation, denoted as $\textbf{e}_{i} \in \mathbb{R}^{d}$. In Fig. \ref{fig_tSNE_visualization}, we visualize $\textbf{e}_{c}$, $\textbf{e}_{p}$ and $\textbf{e}$ of each unseen class and their corresponding visual features. We notice that with our method class semantic word vectors ($\textbf{e}$) and features form clusters, indicating adequate feature-semantic alignment, which is essential for ZSL.

\section{Experiments}

\noindent\textbf{2D Image dataset setup:} For ZSL on 2D images, two well-known datasets are used in this work: (1) The Caltech-UCSD Birds (CUB-200-2011) \cite{WahCUB_200_2011} contains 11,788 images of 200 bird species. We have used the seen-unseen split setting from \cite{Xian_CVPR_2017} with 150 classes as seen data for training and 50 as unseen data for testing. (2) Animals with Attributes (AwA) \cite{Xian_CVPR_2017} dataset consists of 30,745 images of 50 classes with the seen-unseen split of 40 seen classes for training and ten unseen classes for testing.

\noindent\textbf{3D Point cloud dataset setup:} We have used three well-known datasets, ScanObjectNN \cite{scanobjectnn_iccv19}, ModelNet10 and ModelNet40 \cite{wu20153d}. ScanObjectNN dataset consists of 2,902 3D real-world point cloud data of objects of 15 categories. The ModelNet40 dataset contains 12,311 CAD-generated 3D meshes of 40 categories, and ModelNet10 is a part of it. For the Point Cloud datasets, we have used the splits from \cite{cheraghian2021zero}.


\noindent\textbf{Train-Test-Split:} The dataset split used for ZSL divides the data into a seen set and an unseen set. The seen set consists of examples from the classes the model is trained on, while the unseen set contains samples from the classes the model was unknown to. This strategy assesses the model's ability to generalize to unseen classes. In this study, the split configuration from \cite{Xian_CVPR_2017} is used, and the authors emphasized that the choice of dataset split could significantly impact the results and should be selected carefully. The accuracy increases whenever the classes from the test split overlap with the train classes. Therefore, the authors proposed a new split for the train and test splits alongside the standard for all datasets used in this study for ZSL. For 3D datasets, the ModelNet10 and ModelNet40 \cite{wu20153d} datasets are combined, and the 30 classes from the ModelNet40 which are not present in the ModelNet10 are considered as seen classes, and unseen classes are considered from the test set of ModelNet10 following the settings from \cite{cheraghian2021zero}. For the ScanObjectNN \cite{scanobjectnn_iccv19} dataset, we consider 26 classes from the ScanObjectNN dataset that are not overlapped with the ModelNet40 dataset as seen classes, and the overlapped classes are considered seen. 

\noindent\textbf{Language models:} Several LLM models have been used in recent works. Considering the open access and the performance on the NLP domain, we have used the GPT-3.5 turbo\footnote{\url{https://platform.openai.com/docs/api-reference/chat}}  model, which is based on the autoregressive language model GPT-3 from \cite{brown2020language}. The GPT-3.5 turbo model generates additional human-made like text descriptions for each class in datasets. This gives the advantage of having additional annotations of each class without captioning them manually. This description is used along with the available attributes and class names. We also use Word2vec \cite{mikolov2013efficient} to generate the word vector representation of the text descriptions obtained through GPT-3.5. Word2vec learns word embeddings using a neural network from datasets containing billions of words and has millions of words in its vocabulary. And given a text input, it produces a set of vectors representing the words, and it can be used as input alongside the visual features. This gives the advantage of producing high-quality word vectors from the text descriptions using a pre-trained word2vec model to train ZSL models instead of using the attributes vectors produced by manual work alongside the visual features. This work uses the pre-trained 
word2vec model through Gensim \cite{rehurek_lrec} to produce the word vectors. This pre-trained model is trained on a partial Google News Dataset, which contains around 100 billion words. And the trained model contains 300-dimensional vectors for 3 million words and phrases. Passing the obtained class descriptions from GPT-3.5 produces a 300-dim. vector representation for each class.

\begin{table*}[!t] \centering \small
\newcolumntype{C}{>{\centering\arraybackslash}X}
\setlength{\tabcolsep}{4pt}
\scalebox{.9}{
\begin{tabularx}{\textwidth}{clcCCCCC|CCCCC} \hline
\multirow{3}{*}{} & \multirow{3}{*}{Method } & & \multicolumn{5}{c}{\textbf{ModelNet40}} & \multicolumn{5}{c}{\textbf{ScanObjectNN}}\\
\cline{5-13}
{} & {} & {} & ZSL & \multicolumn{4}{c|}{GZSL} & ZSL & \multicolumn{4}{c}{GZSL} \\
 & & Variations & Acc & $\Acc_{s}$ & $\Acc_{u}$ & HM & BC & Acc & $\Acc_{s}$ & $\Acc_{u}$ & HM & BC  \\ 
\hline
& \multirow{3}{*}{DEM \cite{Zhang_2017_CVPR}} & Only Class Name & 17.33 & \textbf{87.69} & 6.35 & 11.84 & 1 & 17.99 & 88.88 & 12.32 & 21.64 & 0 \\
& & Only ChatGPT & \textbf{27.55} & 77.20 & 8.24 & 14.88 & 1 & 13.93 & 88.20 & 4.61 & 8.76 & 0\\
& & Ours & 22.57 & 87.43 & \textbf{10.30} & \textbf{18.43} & \textbf{2} & \textbf{18.48} & \textbf{89.48} & \textbf{14.20} & \textbf{24.51} & \textbf{4} \\
\hline
& \multirow{3}{*}{LATEM \cite{latem-cvpr16}} & Only Class Name & 26.29 & - & - & - & 0 & 11.88 &  & - & - & 0 \\
& & Only ChatGPT & 27.50 &  - & - & - & 0 & 10.61 & - & - & - & 0 \\
& & Ours & \textbf{28.11} &  - & - & - & \textbf{1} & \textbf{13.42} & - & - & - & \textbf{1} \\
\hline
& \multirow{3}{*}{SYNC \cite{Changpinyo_2016_CVPR}} & Only Class Name & 21.17 &  - & - & - & 0 & 17.43 &  & - & - & 0 \\
& & Only ChatGPT & 20.22 &  - & - & - & 0 & 15.90 & - & - & - & 0  \\
& & Ours & \textbf{25.79} &  - & - & - & \textbf{1} & \textbf{18.35} & - & - & - & \textbf{1}  \\
\hline
& \multirow{3}{*}{GDAN \cite{gdan-cvpr19}} & Only Class Name & - & 86.80 & \textbf{2.70} & \textbf{5.23} & \textbf{2} & - & \textbf{88.93} & 16.60 & 27.97 & 1 \\
& & Only ChatGPT & - & 86.87 & 2.25 & 4.38 & 0 & - & 88.18 & 17.33 & 28.97 & 0 \\
& & Ours & - & \textbf{87.07} & 2.36 & 4.60 & 1 & - & 88.50 & \textbf{20.31} & \textbf{33.04} & \textbf{2} \\
\hline
& \multirow{3}{*}{f-CLSWGAN \cite{Xian_2018_CVPR}} & Only Class Name & 18.15 & 88.83 & 1.42 & 2.79 & 0 & 22.51 & 89.17 & 11.83 & 20.91 & 0 \\
& & Only ChatGPT & 19.11 & \textbf{89.33} & 1.10 & 2.17 & 1 & 17.12 & 89.26 & 11.58 & 20.50 & 0 \\
& & Ours & \textbf{26.77} & 88.73 & \textbf{5.90} & \textbf{11.06} & \textbf{3} & \textbf{22.84} & \textbf{89.85} & \textbf{13.01} & \textbf{22.71} & \textbf{4} \\
\hline
& \multirow{3}{*}{TF-VAEGAN \cite{tfvaegan-eccv20}} & Only Class Name & 28.13 & 67.17 & 20.11 & 30.95 & 0 & 28.17 & 75.85 & 25.70 & 38.40 & 0 \\
& & Only ChatGPT & 24.34 & \textbf{76.57} & 18.71 & 30.07 & 1 & 25.86 & \textbf{86.10} & 18.95 & 31.06 & 1 \\
& & Ours & \textbf{30.37} & 73.47 & \textbf{25.45} & \textbf{37.81} & \textbf{3} & \textbf{28.76} & 78.60 & \textbf{26.54} & \textbf{39.68} & \textbf{3} \\
\hline

& \multirow{3}{*}{CADA-VAE \cite{Schonfeld_2019_CVPR}} & Only Class Name & 21.50 & 88.40 & 5.71 & 10.73 & 0 & - & 89.54 & 18.18 & 30.59 & 0 \\
& & Only ChatGPT & 15.61 & 87.03 & 8.21 & 15.01 & 0 & - & 89.17 & 15.43 & 26.31 & 0 \\
& & Ours & \textbf{21.74} & \textbf{88.93} & \textbf{13.16} & \textbf{22.89} & \textbf{4} & - & \textbf{89.92} & \textbf{18.68} & \textbf{30.93} & \textbf{3} \\ 

\hline
\end{tabularx}}
\caption{ZSL and GZSL result on 3D image datasets using PointConv backbone.}
\label{tab:3D_results}
\end{table*}


\begin{table*}[!t] \centering \small
\newcolumntype{C}{>{\centering\arraybackslash}X}
\setlength{\tabcolsep}{4pt}
\scalebox{.9}{
\begin{tabularx}{\textwidth}{clcCCCCC|CCCCC}\hline
\multirow{3}{*}{} & \multirow{3}{*}{Backbone} & & \multicolumn{5}{c}{ModelNet40} & \multicolumn{5}{c}{ScanObjectNN} \\
\cline{4-13}
{} & {} & {} & ZSL & \multicolumn{4}{c|}{GZSL} & ZSL & \multicolumn{4}{c}{GZSL} \\
\cline{4-13}
 & & Variations & Acc & $\Acc_{s}$ & $\Acc_{u}$ & HM & BC & Acc & $\Acc_{s}$ & $\Acc_{u}$ & HM & BC  \\ 
 \hline
& \multirow{3}{*}{PointNet} & Only Class Name & 18.51 & \textbf{56.8} & 15.13 & 23.90 & 1 & 20.48 & 18.06 & \textbf{16.80} & 17.41 & 1 \\ 
& & Only ChatGPT & 23.50 & 46.63 & \textbf{20.80} & \textbf{28.75} & \textbf{2} & 19.78 & 17.92 & 15.80 & 16.80 & 0 \\ 
& & Ours & \textbf{25.64} & 41.53 & 14.44 & 21.43 & 1 & \textbf{21.57} & \textbf{18.77} & 16.25 & \textbf{17.42} & \textbf{3} \\ 
\cline{1-13}
& \multirow{3}{*}{EdgeConv} & Only Class Name & 29.00 & 44.47 & 29.53 & 35.49 & 0 & 23.20 & \textbf{76.03} & 11.40 & 19.87 & 1 \\ 
& & Only ChatGPT & 23.17 & 59.93 & 20.05 & 30.05 & 0 & \textbf{25.41} & 75.95 & 12.06 & 20.82 & 1 \\ 
& & Ours & \textbf{37.86} & \textbf{71.33} & \textbf{29.62} & \textbf{41.86} & \textbf{4} & 24.71 & 75.35 & \textbf{12.79} & \textbf{21.86} & \textbf{2} \\ 
\cline{1-13}
& \multirow{3}{*}{PointConv} & Only Class Name & 28.13 & 67.17 & 20.11 & 30.95 & 0 & 28.17 & 75.85 & 25.70 & 38.40 & 0 \\
& & Only ChatGPT & 24.34 & \textbf{76.57} & 18.71 & 30.95 & 1 & 25.86 & \textbf{86.10} & 18.95 & 31.06 & 1 \\
& & Ours & \textbf{30.37} & 73.47 & \textbf{25.45} & \textbf{37.81} & \textbf{3} & \textbf{28.76} & 78.60 & \textbf{26.54} & \textbf{39.68} & \textbf{3} \\

\hline
\end{tabularx}}
\vspace{.2em}
\caption{ZSL and GZSL performance of TF-VAEGAN \cite{tfvaegan-eccv20} method using our word vector and different backbones.}
\label{table:Performance_different_backbone}
\end{table*}

\noindent\textbf{Evaluation:} 
We have used the top-1 accuracy metric to measure the performance of our models for ZSL, where only the unseen class's information is used to predict an unseen class. For GZSL, the class name is predicted based on both seen and unseen class information. In addition to that, the Harmonic Mean (HM) \cite{Xian_CVPR_2017} of the accuracy of seen and unseen classes in GZSL is also calculated.
$
    HM = \frac{2 \times acc_s \times acc_u}{acc_s + acc_u}  
$
where acc\textsubscript{s} and acc\textsubscript{u} are the top-1 accuracy of the seen and unseen classes, respectively. To understand the better-performing variations in all methods, we have used Borda Count (BC) \cite{saari1995basic}. Each of the variations is only assigned 1 point for ranking highest for each evaluation metric, and the total points are counted to find the best variation.

\noindent\textbf{Validation strategy}: To find the best hyperparameters, we have varied the learning rate, number of epochs, loss weighting factor, and value of $\alpha$. For the 2D datasets with embedding methods, we varied the number of epochs from 1000 to 3000 with the combination of learning rate ranging from 0.001 to 0.00001 and loss weighting factor from 0.01 to 0.001. For the 3D datasets and generative methods, we varied the number of epochs from 100 to 500 with a combination of learning rates ranging from 0.01 to 0.00001. We have varied the value of $\alpha$ from 0.1 to 1.0 with the word vectors. For all the datasets, $\alpha$ values are selected from the set \{0.1, 0.3, 0.5, 0.7, 1.0\} for our proposed method. 

\noindent\textbf{Compared methods:} We compare our work with embedding (DEM \cite{Zhang_2017_CVPR}, LATEM \cite{latem-cvpr16}, SYNC \cite{Changpinyo_2016_CVPR}, GDAN \cite{gdan-cvpr19}) and generative (f-CLSWGAN \cite{Xian_2018_CVPR}, TF-VAEGAN \cite{tfvaegan-eccv20} and CADA-VAE \cite{Schonfeld_2019_CVPR}) methods. These methods mostly reported results on manual attribute vectors for each class. Instead of attributes, we have used word vectors in this paper to eliminate manual annotation. We compare three different variations of experiments for each method: 
\textbf{(1)} Only Classname: Here, the word vector is generated as class semantics using the word2vec model using the class name only. This is the traditional way and serves as a baseline for the ZSL method.
\textbf{(2)} Only ChatGPT: Instead of merely a class name, the text descriptions for each class generated by ChatGPT are used to calculate word vectors using the word2vec model.
\textbf{(3)} Ours: This is our final proposal, where word vectors produced by \textbf{(1)} and \textbf{(2)} are linearly combined to fuse them and produce an improved word vector. Auxilary supervision from ChatGPT helps to improve (noisy) only name-based descriptions.

\noindent\textbf{Implementation details\footnote{Codes and models are available in \url{https://github.com/FHShubho/CGS-ZSL}}:} We have used ChatGPT to generate text descriptions through OpenAI's API. The text descriptions, as well as the class names, are passed through a pre-trained 
word2vec model using Gensim \cite{rehurek_lrec} to generate word vectors from class names and ChatGPT text descriptions. The word vectors are then combined to create our variation of the word vector. These word vector variations are used with the visual features extracted from the datasets' 2D images and 3D point clouds. For the 2D images, ResNet101 is used to extract the visual features similar to \cite{Xian_CVPR_2017}, which produces visual features of 2048 dimensions.  For the 3D point clouds, we used PointNet \cite{qi2017pointnet}, PointConv \cite{8954200}, and EdgeConv \cite{Article24} to extract visual features from the 3D point clouds where the first two produce visual features of 1024 dimensions and the later one of 2048 dimensions. 


\subsection{Performance on 2D Image datasets} 
The performances are measured for ZSL and GZSL settings and shown in Table \ref{tab:2D_results}. The following observations can be made from the results. \textbf{(1)} For all the methods, the performance consistently improves while using the image features with the combined word vector of the class names and descriptions. \textbf{(2)} The results are better for the AwA datasets than those on CUB and 3D datasets. One reason is that the animal classes in the AwA dataset are more distinguishable in terms of visual and descriptive features. In contrast, the CUB dataset contains only bird classes with similar physical and descriptive features. So describing the birds in natural language will be more similar in words, and it will also be reflected in the word vectors generated from the descriptions. \textbf{(3)} Using only ChatGPT-based word vectors from the description produces better results than only class name-based ones. For AwA datasets, using word vectors for descriptions gives better ZSL results than the other two variations. But overall, our proposed combined word vector of class name and description holds top BC score on both datasets over all methods. It tells us that our proposed method can improve the quality of class semantics for ZSL and GZSL.

\subsection{Performance on 3D Point cloud datasets} 

We extract 3D point cloud features from the PointConv backbone and apply and align them with the word vectors generated using the Word2vec model. Table \ref{tab:3D_results} we report 3D ZSL results. Here are our observations: 
\textbf{(1)} Our proposed combined word vector of class name and description improves performance across all the methods. For both ZSL and GZSL settings, accuracy is higher than using word vectors of only labels or only ChatGPT descriptions. The BC score of our combined vector of class name and description also affirms that it improves performance for all the methods.
\textbf{(2)} For embedding-based methods like DEM and GDAN, the performance of only ChatGPT description word vector is better than the word vector of only the class name. The combined word vector of the class name and description beats the previous ones. For the generative models like TF-VAEGAN and CADA-VAE, the performance with the word vector of the class name is better than with the word vector of the ChatGPT description. And again, the performance of our proposed combined word vector is better than the previous ones. 
\textbf{(3)} The methods used for 2D ZSL can be used for 3D point clouds by using the extracted features of the 3D point cloud as input in place of the features extracted from the images of the 2D dataset. And the results show that the methods from 2D ZSL also perform well for the 3D point cloud datasets. 
\textbf{(4)} Using our proposed combined word vector representation consistently improved the ZSL performance and the unseen class in the GZSL setting.
\textbf{(5)} For the ScanObjectNN dataset case, the training is done on synthetic ModelNet40 objects, but testing is done on real-word-scanned (not synthetic) ScanObjectNN objects. This is a complex setup because the domain of train and testing is different. Working consistently in this challenging setting proves the robustness of our approach.


\begin{figure}[!t]
\centering
\includegraphics[width=.46\textwidth]{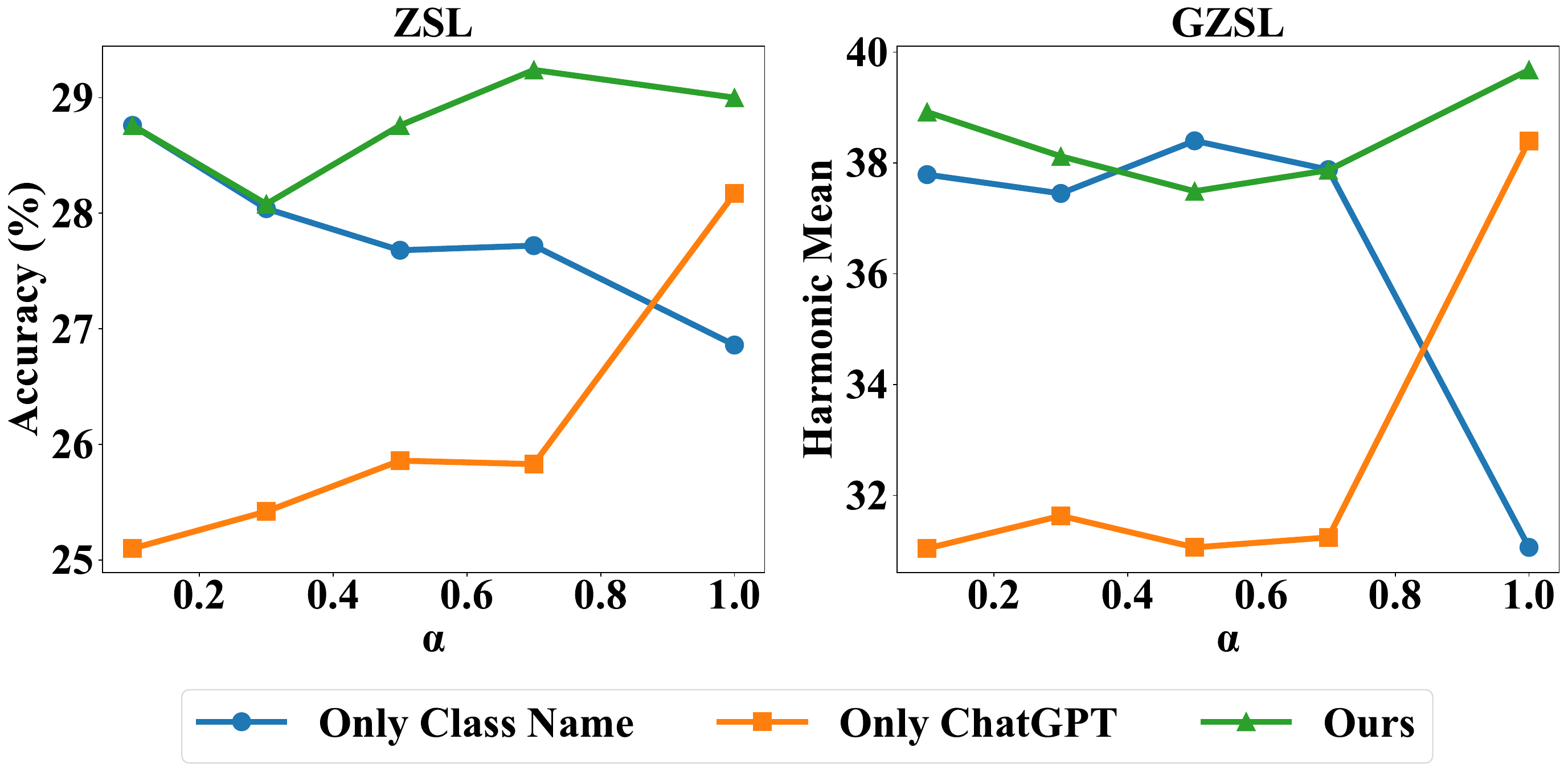}  
\caption{ZSL and GZSL for different values of $\alpha$ for TF-VAEGAN \cite{tfvaegan-eccv20} on ScanObjectNN dataset}
\label{fig_performance_over_different_alpha}
\end{figure}
\subsection{Ablation studies}

\noindent\textbf{Different 3D backbones:} We have experimented with three different backbones for 3D datasets,  PointNet \cite{qi2017pointnet}, PointConv \cite{8954200}, and EdgeConv \cite{Article24}. These three backbones extract point cloud features from the 3D object, later used as the input with all the variants of the word vector representations. As shown in Table \ref{table:Performance_different_backbone}, the features extracted through the PointConv backbone yield better performance than PointNet and PointConv backbones. We experimented with all the backbones for all seven methods and observed that the features from PointConv showed better performance. We notice that using the word vectors from ChatGPT and our representation of word vectors improves performance consistently across all the backbones.

\noindent\textbf{Varing $\alpha$:} While combining the word vectors generated using the class name and text description from ChatGPT, we varied the value of $\alpha$. We experimented with the $\alpha$ values of 0.1, 0.3, 0.5, 0.7, and 1.0. Fig. \ref{fig_performance_over_different_alpha} shows the accuracy of ZSL and Harmonic Mean in the GZSL setting over varying values of $\alpha$. We can observe that increasing the value of $\alpha$ yield performance improvements for ZSL and GZSL with our vector representation. 
We can also observe that the word vector of the text representation from the ChatGPT also increases with the increasing values of $\alpha$. Still, the performance with the word vector of only class name decreases as the value of $\alpha$ increases.



\section{Conclusion}
This paper investigates the possible use of ChatGPT to improve class semantics for zero-shot learning tasks. ChatGPT-based word vectors fused with traditional class name-based vectors can achieve better ZSL and GZSL performance on 2D image and 3D point cloud recognition tasks. In experiments, we apply our proposal to multiple embedding-based and generative ZSL methods and notice that our technique consistently improves those methods' existing performance. We perform extensive experiments on 2D image and 3D point cloud datasets. It tells that ChatGPT could be a suitable annotation tool that can provide automatic and less noisy information without manual labor. In the future, we will investigate the use of ChatGPT-guided embedding vectors for object detection and segmentation tasks.

{
\bibliographystyle{IEEEtran}
\bibliography{egbib}
}

\end{document}

%% file: macros.tex
\DeclareMathOperator{\Acc}{Acc}

\newcommand{\be}{\mathbf{e}}

\newcommand{\sE}{\mathcal{E}}

\newcommand{\sY}{\mathcal{Y}}

\newcommand{\bbR}{\mathbb{R}}

